\pdfoutput=1
\documentclass[11pt]{article}

\usepackage{ACL2023}

\usepackage{times}
\usepackage{latexsym}

\usepackage[T1]{fontenc}
\usepackage[utf8]{inputenc}

\usepackage{microtype}

\usepackage{inconsolata}
\usepackage{booktabs}

\usepackage{wrapfig}
\usepackage{hyperref}
\usepackage{graphicx}
\usepackage{multirow}
\usepackage{hyperref}
\usepackage{bm}
\usepackage{xcolor}
\usepackage{color}

\usepackage{stmaryrd}
\usepackage{fixltx2e}
\usepackage{xparse}
\usepackage{amssymb}
\usepackage{amsfonts}
\usepackage{soul}

\usepackage{mathtools}
\usepackage{comment}

\DeclareGraphicsExtensions{.png,.pdf}
\label{packs}

\NewDocumentCommand{\teal}{ }{\textcolor{teal}}

\NewDocumentCommand{\heng}{ mO{} }{\textcolor{red}{\textsuperscript{\textit{Heng}}\textsf{\textbf{\small[#1]}}}}

\NewDocumentCommand{\chenkai}{ mO{} }{\textcolor{blue}{\textsuperscript{\textit{Chenkai}}\textsf{\textbf{\small[#1]}}}}

\NewDocumentCommand{\cheng}{ mO{} }{\textcolor{purple}{\textsuperscript{\textit{Cheng}}\textsf{\textbf{\small[#1]}}}}

\NewDocumentCommand{\jinning}{ mO{} }{\textcolor{teal}{\textsuperscript{\textit{Tie}}\textsf{\textbf{\small[#1]}}}}

\NewDocumentCommand{\ken}{ mO{} }{\textcolor{orange}{\textsuperscript{\textit{Ken}}\textsf{\textbf{\small[#1]}}}}

\newcommand\ct[1]{~\cite{#1}}
\newcommand\fu[1]{\footnote{\url{#1}}}

\newcommand\blfootnote[1]{%
  \begingroup
  \renewcommand\thefootnote{}\footnote{#1}%
  \addtocounter{footnote}{-1}%
  \endgroup
}
\title{Measuring the Effect of Influential Messages on Varying Personas}
\author{
Chenkai Sun\textsuperscript{$\spadesuit$}, Jinning Li\textsuperscript{$\spadesuit$}, Hou Pong Chan\textsuperscript{$\heartsuit$}, ChengXiang Zhai\textsuperscript{$\spadesuit$}, and Heng Ji\textsuperscript{$\spadesuit$} \\
\textsuperscript{$\spadesuit$}University of Illinois Urbana-Champaign \\
\textsuperscript{$\heartsuit$}Faculty of Science and Technology, University of Macau \\
\textsuperscript{$\spadesuit$}\texttt{\{chenkai5, jinning4, czhai, hengji\}@illinois.edu}  \\
\textsuperscript{$\heartsuit$}\texttt{hpchan@um.edu.mo}
  \\}

\begin{document}
\maketitle

\begin{abstract}

Predicting how a user responds to news events enables important applications such as allowing intelligent agents or content producers to estimate the effect on different communities and revise unreleased messages to prevent unexpected bad outcomes such as social conflict and moral injury. We present a new task, Response Forecasting on Personas for News Media, to estimate the response a persona (characterizing an individual or a group) might have upon seeing a news message. Compared to the previous efforts which only predict generic comments to news, the proposed task not only introduces personalization in the modeling but also predicts the sentiment polarity and intensity of each response. This enables more accurate and comprehensive inference on the mental state of the persona. Meanwhile, the generated sentiment dimensions make the evaluation and application more reliable. We create the first benchmark dataset, which consists of 13,357 responses to 3,847 news headlines from Twitter. We further evaluate the SOTA neural language models with our dataset. The empirical results suggest that the included persona attributes are helpful for the performance of all response dimensions. Our analysis shows that the best-performing models are capable of predicting responses that are consistent with the personas, and as a byproduct, the task formulation also enables many interesting applications in the analysis of social network groups and their opinions, such as the discovery of extreme opinion groups. \blfootnote{Code Repository: \url{https://github.com/chenkaisun/response_forecasting}}

\end{abstract}

\begin{figure}
	\vskip 0.2in
	\centering
	\includegraphics[width=1\linewidth]{"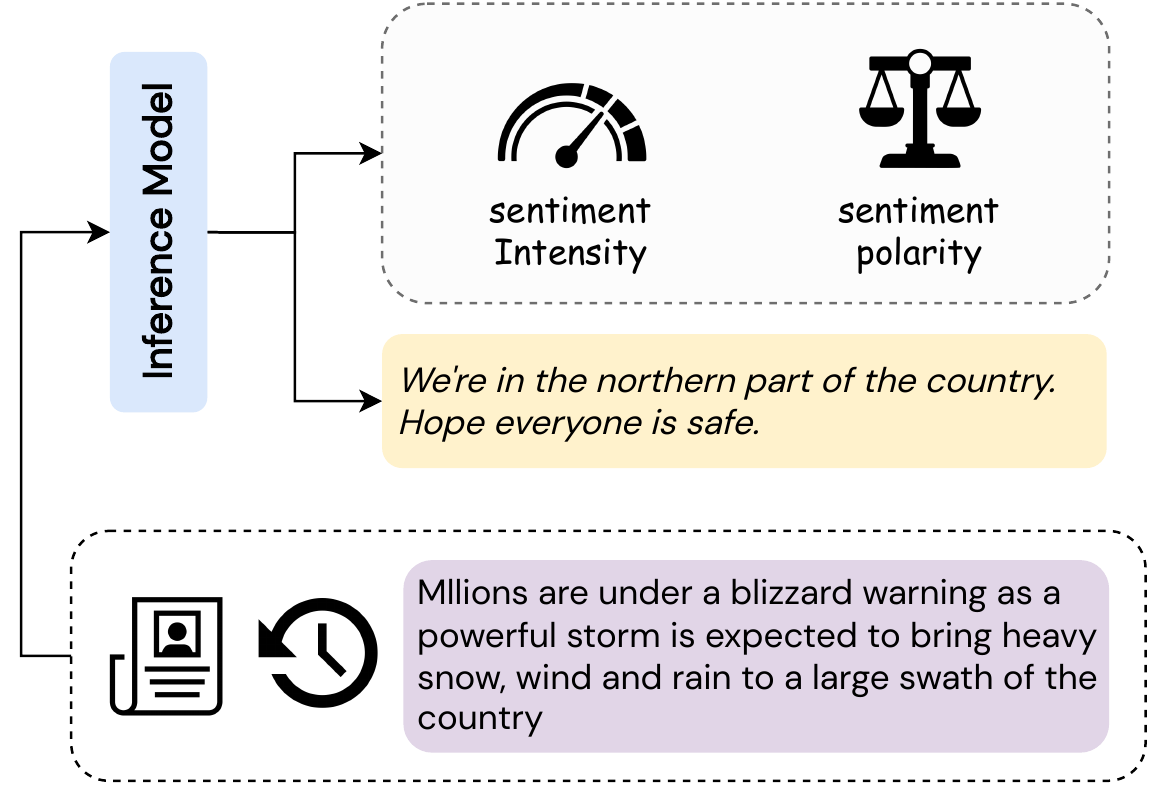"}
	\caption{An example illustrating the task. The input consists of persona attributes (e.g., historical activities and profile) and a news message. The model is asked to predict response in multiple dimensions.}
	\label{fig:example}
	\vskip -0.2in
\end{figure}

\section{Introduction}
\label{sec:intro}
% flooding of 
To prevent the flooding of misinformation and hate speech on the internet, a great amount of progress has been made toward identifying and filtering such content on social media using machine learning models~\cite{misinfo1,misinfo2,hatespeech1,hatespeech2}. While directly creating message-level labels is a natural way to address the issue, it is equally important to measure the influence of the message on different viewers as a way to decide how to manage the publication of the messages. 
% a ``nutritional label'' to 
% prior to 

Existing efforts~\cite{rdistr1,rdistr2,newcommentgen,artzi2012predicting} have made steps toward predicting population-level news response (e.g., predicting the most likely response to a news message), but neglected the importance of personas in measuring influence. According to Individual Differences Theory~\cite{idt}, which proposes that individuals respond differently to the mass media according to their psychological needs, the same message can impact different population groups/personas in different ways. For example, a message claiming the honor of sacrificing others' lives for a religious goal might %is free of hate speech and misinformation, but it can nonetheless 
agitate people who are prone to agreeing with such messages. It is therefore essential to consider personalization when inferring viewers' responses. 

On the other hand, the previous approaches that predict text-level responses~\cite{newcommentgen, prp1,lu2022partner} have only used generation metrics for automatic evaluation, yet %a person 
%who is happy and supportive of the news message 
%can express such 
the same sentiment can be expressed in a multitude of ways, and text alignment metrics like BLEU\ct{bleu} and ROUGE\ct{rouge} do not credit cases where the sentiments match but semantics do not 
%don't\ken{avoid using don't, use do not} 
align well. As a result, it is crucial to evaluate the sentiment dimensions of user responses. %\ken{this paragraph is too long, consider break it into two. }
% On the other hand, there are existing datasets on generating from for answering   user comment to response prediction have neglected the multiple d only comment  a single aspect of an individual response (e.g., emotion, tex)
% indicating the healthiness of the text (at the viewer's end)

% therefore

% Existing work (cite distr) involves predicting population-level respobnse distribution, the works have been neglected. Therefore it is crucial to take into account of how modeling how different personas would respond differently to . Other works have defined the response as text generation, yet the utterance doesn't reflect sentiment accurately (satirizng or not, and how agitated is the user).
% Specifically, we argue that 
% , since a message free of hate speech and misinformation can still have influential effects on the viewers
 % prior to publication
% certain religious 
% According to Individual Differences Theory (ct), ... .  
% on viewers' effect at the speaker's end before even releasing. Moreover, misinformation
 % (which can be human/machine-generated)
% through filtering 

% prop and misinfo, Nowadays, social media
% content producer
% limited
% influence

% existing task on nonpersonalized sent pred
% AGPC Individual Differences Theory
% Despite

% existing tasks and data
% deficiencies
% \ken{better to highlight that the outputs are personalized, e.g., predicting personalized responses for viewers}
 % \ken{You should mention that the input of this task is persona attributes in somewhere.}
  % \heng{explain what is in persona information}
We propose Response Forecasting on Personas for News Media, a task for measuring the influence of news media messages on viewers by predicting viewers' responses. In particular, the input consists of the news message and persona information (e.g., user profile and history in our dataset), and we define response in terms of sentiment polarity, sentiment intensity, and textual response.  While we include three categories in this work, many other interesting aspects can also be defined (e.g., change of attitude toward real-world entities) and we leave them to future work. Studying the problem of forecasting individual viewers' responses allows the creation of tools to assist analysts and online content producers to estimate the potential impact of messages on different communities, and sheds light on new applications such as automatically re-writing a message/email to achieve a communication goal (e.g., to obtain a positive response from the receiver). Furthermore, this new task also helps to understand associations between user attributes and emotional responses.

% \ken{you use both past tense and present tense in the introduction section, need to be consistent}
To construct a test bed for this task, we collect a dataset from Twitter consisting of 13,357 labeled responses to 3,847 news headlines from Twitter. Using the corpus, we examine how state-of-the-art neural models work in our task. We find that the models can predict responses with reasonable accuracy yet still have a large room for improvement. We also find that the best-performing models are capable of predicting responses that are consistent with the personas, indicating that the models may be used for many exciting applications such as the discovery of groups with different opinions.

{\renewcommand{\arraystretch}{1} 
\begin{table}[t]
\begin{small}
	\centering
    \begin{tabular}{l|ccc}
   \toprule
Split & Train & Dev. & Test \\\midrule
\# Samples & 10,977  & 1,341 & 1,039 \\
\# Headlines & 3,561 & 1,065 & 843 \\
\# Users & 7,243 & 1,206  & 961 \\
% \% Empty Profiles  &  &  &  \\
% \% Empty History &  &  &  \\
Avg \# Profile Tokens &10.75  & 11.02 & 10.50 \\
Avg \# Response Tokens & 12.33 & 12.2 & 11.87 \\
Avg \# Headline Tokens &19.79  &  19.82& 19.72 \\\bottomrule 
\end{tabular}
    
    \caption{Summary statistics for the dataset.}
	\label{tab:data_stats}
% (\%) 
	\vskip -0.3 in
\end{small}
\end{table}
}

% Avg \# Historical Tweet Tokens  & 14.92 &14.90  &14.87  \\

% {\renewcommand{\arraystretch}{1} 
% \begin{table}[t]
% 	\centering

%     \begin{tabular}{l|lll}
%    \hline
% Name & Train & Dev. & Test \\\hline
% \# Samples &  &  &  \\
% \# Unique Headlines &  &  &  \\
% \# Users &  &  &  \\
% \% Empty Profiles  &  &  &  \\
% \% Empty History &  &  &  \\
% Avg \# history tokens  &  &  &  \\
% Avg \# profile tokens &  &  &  \\\hline 
% \end{tabular}
    
%     \caption{Summary statistics for the Dataset. The Train Dataset is labeled by existing model. (Note Intensity distr)}
% 	\label{tab:topic_dist}
% % (\%) 
% \end{table}
% }

% \usepackage[dvipsnames]{xcolor}

\begin{table*}[]
\begin{small}
% \centering
\centering
\begin{tabular}{lllllllllll}
\toprule
 & \multicolumn{6}{c}{\textbf{Textual Response}} & \multicolumn{2}{c}{\textbf{$\phi_{int}$}} & \multicolumn{2}{c}{\textbf{$\phi_{p}$}} \\
Name & \textcolor{violet}{BLEU}& \textcolor{violet}{BScore} &\textcolor{violet}{Meteor} & \textcolor{violet}{R-1} &\textcolor{violet}{R-L} & \multicolumn{1}{c|}{\textcolor{violet}{Avg. Len}} &\multicolumn{1}{c}{\textcolor{orange}{$r_s$}} & \multicolumn{1}{c|}{\textcolor{orange}{$r$}} & \teal{MiF1} & \teal{MaF1} \\ \midrule
% Random &  &  &  &  &  & \multicolumn{1}{c|}{} &  & \multicolumn{1}{c|}{} &  &  \\
% Majority &  &  &  &  &  & \multicolumn{1}{c|}{} &  & \multicolumn{1}{c|}{} &  &  \\
% GPT &  &  &  &  &  & \multicolumn{1}{c|}{} &  & \multicolumn{1}{c|}{} &  &  \\

Majority & - & - & - & - & - & \multicolumn{1}{c|}{-} & - & \multicolumn{1}{c|}{-} & 43.41 & 20.18 \\
Random & - & - & - & - & - & \multicolumn{1}{c|}{-} & 0.62 & \multicolumn{1}{c|}{0.41} & 35.51 & 30.55\\
GPT2 & 1.59 & -5.78 & 3.36 & 6.50 & 1.90 & \multicolumn{1}{c|}{9.64} & 50.34 & \multicolumn{1}{c|}{49.78} & 60.25 & 56.85 \\
T5 & 6.95 & -5.71 & 5.98 & \textbf{10.40} & \textbf{2.70} & \multicolumn{1}{c|}{18.87} & 50.06 & \multicolumn{1}{c|}{49.26} & 63.72 & 57.85 \\
BART & \textbf{8.17} & \textbf{-5.67} & \textbf{6.09} & 9.90 & 2.50 & \multicolumn{1}{c|}{\textbf{21.05}} & \textbf{62.03} & \multicolumn{1}{c|}{\textbf{61.82}} & \textbf{67.85} & \textbf{63.23} \\
BART w/o Profile & 7.30 & -5.70 & 5.91 & 10.00 & 2.50 & \multicolumn{1}{c|}{19.47} & 57.95 & \multicolumn{1}{c|}{58.20} & 67.28 & 62.26 \\
BART w/o History & 5.24 & -5.88 & 4.41 & 7.70 & 1.50 & \multicolumn{1}{c|}{18.62} & 48.80 & \multicolumn{1}{c|}{48.63} & 59.00 & 53.29 \\
BART w/o Both & 3.90 & -5.92 & 4.00 & 7.90 & 1.80 & \multicolumn{1}{c|}{15.73} & 45.28 & \multicolumn{1}{c|}{44.75} & 61.41 & 46.01 \\

 \bottomrule
% Human & \multicolumn{1}{l}{100} & \multicolumn{1}{l}{100} & \multicolumn{1}{l}{100} & \multicolumn{1}{l}{100} & \multicolumn{1}{l}{100} & \multicolumn{1}{l|}{100} & \multicolumn{1}{l}{100} & \multicolumn{1}{l|}{100} & \multicolumn{1}{l}{100} & \multicolumn{1}{l}{100} \\ \hline

\end{tabular}
\caption{Response forecasting results above show that the state-of-the-art models can predict responses with reasonable performance. The best overall performance is bolded.}
\label{tab:main}
\end{small}
\end{table*}

\section{Dataset Collection}
\label{sec:dataset}
% \begin{table}
% 	\centering
% 	\begin{tabular}{lllll}
		
% 		\toprule
% 		Setting &Anno.& \#Doc.& \#Sent. & \#Mention   \\
% 		\midrule
% 		Train &Distant&    80&6565&10318\\
% 		Dev &Human&20&520&1118\\
% 		Test&Human&            20&663&1295 \\
% 		\bottomrule
% 	\end{tabular}
% 	\caption{Dataset Statistics for CHEMET}
% 	\label{datastats}
% \end{table}
In this section, we describe how we construct data from Twitter. Specifically, we used Twitter API\footnote{\url{developer.twitter.com/en/docs/twitter-api}} to crawl news headlines and comments below each headline from CNN Breaking News\fu{twitter.com/cnnbrk}, which is one of the most popular news accounts on Twitter.

% . After filtering, we in total collect $6685$ headlines and $54897$ replies posted between 2017/01/01 and 2019/01/01. In this paper, we focus on monolingual setting so we filter out non-English data.

\noindent\textbf{Preprocess}. 
We collected news headlines and corresponding comments from CNN Breaking News between January 2017 and January 2019 and removed the comments that are over 50 tokens to avoid spamming. We stripped away HTML syntax tokens and normalized user reference with special tokens ``@user''.

% The non-English comments are also removed since our dataset focuses on English. Lastly,

% Observing that spamming comments almost never receive any vote,  w
% url links, emails, and digits in our corpus with unique tokens “url”, “email” and “digit”. on all topics excluding politics, as we later 

% we stripped away potential markdown and Html syntax tokens and replaced all forms of url links, emails, and digits in our corpus with unique tokens “url”, “email” and “digit” respectively.

% After filtering, we in total collect $6685$ headlines and $54897$ replies posted between 2017/01/01 and 2019/01/01.

% \subsection{Headline-Comment Relevance}
% \label{hc-relevance}
% To ensure the quality of user comments (e.g., excluding non-relevant comments or spam),  we randomly sampled 300 headline-comment pairs from our data and asked annotators to judge whether the comment is responding to the headline. The annotator can select ``yes'', ``no'' or ``uncertain'' as the answer. The annotators are students from the computer science department who have passed an initial small quiz on the same task.

\subsection{Persona Data}
% We refer to the users who produce the comments as responders. To describe responders, we collected different classes of persona attributes from Twitter. We classify persona attributes into being either observable or latent. Specifically, observable persona attributes are attributes for which user describes themselves publicly, and latent attributes are the ones that the user typically doesn't publicly share (or know) and need to be inferred (e.g., moral value and ideology). For observable attributes, we collected (2) user self-description; (3) user historical posts, which are tweets written directly by the user. For user historical posts, to ensure that future posting activities are not included when predicting the comment, we collect the historical posts prior to the earliest data sample in our dataset for a particular user. 
We categorize the users who post comments as responders. To describe responders, we gathered various persona attributes from Twitter, including (1) User Profile, which is a short paragraph describing the user, and (2) User History, which are tweets written directly by the user. We consider persona as a representation of an individual or a community that characterizes interests and beliefs. User profiles and history serve as effective indicators of persona, as they reveal such information well. Since users' behavior is generally influenced by their personas, we can potentially infer personas by analyzing data that reflects their behavior. Additionally, studying historical tweets helps us understand users' communication styles. To ensure that future posting activities are not included when predicting the comment, we collect the historical posts prior to the earliest data sample in our dataset for each individual user.

% We include statistics on user history
% (4) social network, which contains user-news reply edges and user-user following edge. To alleviate sparsity in social network, we augment the graph with users in the following list of each of responders. 

% \noindent\textbf{User Profile and Historical Posts}

% <motivation
% Users’ comment histories can often signal their personal preferences toward topics or even texting habits as shown in Table 1, thus it is beneficial to collect these histories.
% We obtained a user’s comment histories by querying the Pushshift Reddit API
% <desc
% \noindent\textbf{Social Network}

% \noindent\textbf{Analysis}
% TODO. An analysis is shown in Table X. Graph Profile Historical Posts Analysis, 

% for sentiment polarity, we use the state-of-the-art model\footnote{\url{https://huggingface.co/cardiffnlp/twitter-roberta-base-sentiment-latest}} based on the TweetEval task~\cite{tweeteval} dataset, reaching over 73\% accuracy; for sentiment intensity, 
 % \heng{what preprocessing did you do?}
 
\subsection{Annotation}
We obtained 14k headline and comment pairs from preprocessing. In the annotation stage, we collect labels for sentiment intensity and polarity of comments based on the context of the headline. For the 10k training instances, we produce automatic labels using deep-learning models trained on existing message-level datasets. More specifically, we train a Deberta-based model\ct{deberta} using data from SemEval-2018 Task 1\footnote{\url{https://competitions.codalab.org/competitions/17751}} \ct{mohammad2018semeval}, reaching over 85\% Pearson correlation. We then proceed to use crowd-sourcing to annotate the remaining 2k samples as our evaluation set.

\noindent\textbf{Task Setup}. The annotation for the evaluation set is performed using the Amazon Mechanical Turk (MTurk) crowd-sourcing platform. The workers were each asked to annotate a headline and comment pair with three workers assigned to each data sample. During the annotation, the annotator is asked to select the  sentiment polarity label and the intensity of the sentiment based on their understanding of the input. The workers select  positive, negative, or neutral for the sentiment polarity label and select on the integer scale of 0 to 3 for intensity. 415 workers participated in this task in total and all annotators are paid a fair wage above the federal minimum.

% To give the annotator more freedom on granularity, we design the intensity to be on a scale of 1-9. The score ranges 1-3, 4-6, and 7-9 are respectively mapped to \textit{low}, \textit{medium}, and \textit{high}.

% The interface layout is shown in Figure X. 
\noindent\textbf{Quality Control}. To ensure the quality of annotation, we allowed only the workers who have at least 95\% approval rate and have had at least 5,000 hits approved to access our tasks. We further removed workers who have a <70\% accuracy in the first 30 annotations and discarded the assignments that have completion time deviated from the expected average largely. We used majority voting to determine the final labels: if at least two annotators agreed on a label, we chose it as the final label. The resulting annotated samples achieve an inter-annotator agreement accuracy of 81.3\%. We show the statistics of the dataset in Table~\ref{tab:data_stats}.

\section{Response Forecasting on Personas for News Media}
\label{sec:experiment}
\subsection{Task Formulation}
% ====work in progress==== (optional)
% social network, 
In this task, we aim to predict sentiment polarity, sentiment intensity, and textual response from an individual when the individual sees a message on news media. Formally, given persona $\mathcal{P}$ (represented by profile, or historical posts), and a source message $\mathcal{M}$, the task is to predict the persona's sentiment polarity $\phi_p$ (i.e., \textit{Positive}, \textit{Negative}, \textit{Neutral}) and sentiment intensity $\phi_{int}$ (i.e.,  in the scale of 0 to 3), and textual expression $t$. Our goal is to encode $\mathcal{P}$ and produce $\phi_p$, $\phi_{int}$, and $t$ at decoding time. We formulate the task as a conditional generation problem and use the following maximum-likelihood objective to train a generative model:

	\vskip -0.2in

\begin{gather*}
	\sum_i^N \log p(O_i|O_{<i-1}, \mathcal{P})
		% \end{split}
\end{gather*} 
where $O$ is the output string concatenating $\phi_p$, $\phi_{int}$, and $t$ with special separator tokens.

% Our input is formulated as 
% we formulate the problem as conditional generation and, during training, we maximize the standard language modeling objective

% 	\vskip -0.2in
% \begin{gather*}
% 	% \begin{split}
% 		p(\text{Completion of Script}|\mathcal{H, P, G})  \\  = \prod_i p(\text{Script}_i|\text{Script}_{<i},\mathcal{P, G})
% 		% \end{split}
% \end{gather*} 

% \begin{gather*}
% 	% \begin{split}
% 		\mathcal{I}=\{\mathcal{P} ,\mathcal{M}\}\notag\\  =\{\texttt{<BOS>}  m_1, m_2, ..., m_M \notag\\\texttt{<SE>} s \texttt{<UP>} u_1, u_2, ..., u_U \notag\\ \texttt{<HIST>} h_1, h_2, ..., h_H  \texttt{<EOS>}\notag \} 
% 		% \end{split}
% \end{gather*} 

\subsection{Experimental Setup}

% We test the ability of large-scale language models to generate inferences for unseen news headlines using conditional generation (Sutskever et al., 2014; Rush et al., 2015). We use topic- and dimensionbased special tokens to control generation of belief frames for T5 encoder-decoder (Raffel et al., 2020) and GPT-2 decoder-only models (Radford et al., 2019). Both models are based on a transformer architecture (Vaswani et al., 2017), consisting of transformer blocks with self-attention and feedforward layers as well as layer normalization.

For deep learning-based text generators, we fine-tune decoder-only text generator GPT2\ct{gpt2} as well as two Encoder-Decoder models T5\ct{t5} and BART\ct{bart}. Greedy decoding is used for all the models during training. We further perform ablation on the best-performing model by removing different user attributes. We further include two naive baselines, \textit{Random} and \textit{Majority}, for sentiment dimensions, where each prediction follows either the majority label or a random label. Our neural models are  implemented  using Pytorch~\cite{pytorch}  and  Huggingface Transformers~\cite{huggingface}. The reproducibility and hyperparameter details can be found in Appendix Table~\ref{tab:hparam}.
% We further include three simple baselines, \textit{Random} retrieval of response under the same headline, and using \textit{SBERT}\ct{sbert} to retrieve the response from the closest persona under the same headline. %For SBERT, we use the concatenation of user id, profile, and most recent as key 

% \subsection{Implementation and Training Detail}

% Our  model  is  implemented  using Pytorch~\cite{pytorch}  and  Huggingface Transformers~\cite{huggingface} with BART-base as a base generator. Our model is trained on a single 32GB NVIDIA Tesla V100 GPU. The reproducibility and hyperparameter details can be found in the appendix.

% \subsubsection{Text Encoding}

% \noindent\textbf{Bert} 

% \noindent\textbf{Code Generation} 

% \subsubsection{Social Context Encoding}

% \noindent\textbf{Node2Vec} 

% \noindent\textbf{GraphSage}

% \subsubsection{Speaker Style Modeling}

% \noindent\textbf{Speaker Pretrain}

% \noindent\textbf{Concatenation of Historical Posts} 

% (1/2/3) ngram overlap metric BERTScore~\cite{bertscore}, BERTScore is a model-based metric to measure the similarity between generated and references.  Micro and Macro-

% \begin{figure}
% 	\vskip 0.2in
% 	\centering
% 	\includegraphics[width=0.9\linewidth]{"human_eval.png"}
% 	\caption{A plot showing human evaluation results.}
% 	\label{fig:human_eval}
% 	\vskip -0.2in
% \end{figure}

{\renewcommand{\arraystretch}{1} 
\begin{table}[t]
	\centering
    \begin{tabular}{l|ccc}
   \toprule
Model & Persona& Label & Context \\\midrule
GPT2 &3.18  & 3.84 & 2.84 \\
T5  & 3.68 & 4.23 & 3.57 \\
BART& \textbf{4.35} &\textbf{4.42}  & \textbf{3.99} \\
% \% Empty Profiles  &  &  &  \\
% \% Empty History &  &  &  \\
\bottomrule
\end{tabular}	
    \caption{The table shows human evaluation results based on three consistency measures, supporting the automatic evaluation findings.}
	\label{tab:human_eval}
% (\%) 
\vskip -0.2in
\end{table}
}

\subsubsection{Evaluation Metrics}
\noindent\textbf{Automatic}. We use BARTScore~\cite{bartscore}, BLEU~\cite{bleu} , METEOR~\cite{meteor}, and ROUGE~\cite{rouge} to evaluate textual response generation performance. Note that BARTScore computes the log-likelihood of producing the reference text given the generated text using a BART model pretrained on ParaBank2\footnote{\url{https://github.com/neulab/BARTScore}}. Furthermore, we use Pearson and Spearman correlation to evaluate sentiment intensity, and F1 to evaluate sentiment polarity. 

\noindent\textbf{Manual}. We conduct human evaluation to measure the consistency of the generated outputs from those models. We define three types of consistency metrics: (1) \textit{persona consistency}: whether the output reflects the persona's characteristics, (2) \textit{label consistency}: whether the response text and sentiment are consistent with each other, (3) and \textit{context consistency}: whether the output is responding to the input news headline. We randomly select 10 personas with distinct characteristics (i.e., the writing style/interest/profession do not clearly overlap) and 10 news headlines from distinct topics, and consequently generate 100 responses using each model. The samples are distributed to 5 raters who score each output based on our metrics. The raters are master students who passed a small quiz of 20 samples with at least 80\% accuracy. We additionally make sure that each rater is familiar with the persona information (e.g., profile and history) before starting to work on the task.
% from Section~\ref{hc-relevance}

\subsection{Results}

\noindent\textbf{Automatic Evaluation}.
\label{sec:autoeval}Across the metrics in Table~\ref{tab:main}, we can see that BART provides us with the highest quality response predictions on both sentiment and text levels. As expected, the performance of simple baselines is relatively low compared to other models, showing that the dataset does not have a class imbalance issue. While the automatic generation scores are generally low (i.e., words do not align well), the sentiment prediction scores are much higher in scale, demonstrating the importance of sentiment scoring to make a fair judgment of the result; the model needs to be credited for correctly predicting the latent sentiment even if it does not utter the exact sentence. Finally, we ablate user attribute features one by one. As shown in the table, not only both features included are effective for the task, but they are also complementary of each other.

% We further train two variant models Bart text-only, which is trained solely on response text generation, and Bart sentiment-only, which is trained on sentiment prediction only. We find that the original multitasking version outperforms both variants, demonstrating that multi-tasking may be a more suitable way of training for this task. 

% The BertScore is close since it's based soft-matching between 
%   Table~\ref{table:test}
% The test set performance can be seen in the Appendix.
%  \heng{shouldn't you put test set results in the main text?}
% Moreover, while CRA achieves the best in Finance and Food categories on all metrics, it is not the case in Vehicle. This shows the level of effectiveness also depends on the domain. 
% \noindent \textbf{Results} The test set results are shown in Table~\ref{tab:test}.

\noindent\textbf{Human Evaluation}.
\label{sec:humaneval} The results from human judgments (Table~\ref{tab:human_eval}) in general support the automatic evaluation findings. Among all three models, our approach with BART reaches the highest on all metrics, showing it can generate responses of better quality than others. The difference between models on Label Consistency is noticeably lower than other metrics, and the number suggests that pretrained language models are capable of producing sentiment labels consistent with the textual expression. On the other hand, we find that BART can produce responses more consistent with the controllable variables than GPT2, which might be attributed to its denoising pretraining (e.g., it adapts better to different modeling formats). In fact, the outputs show that GPT2 hallucinates more often than other models. %, perhaps due to its causal language model pre-training.
% , as GPT2 is pretrained on causal language modeling on natural language

% \subsection{Human Evaluation}
% the result from human judgments（bar chat） in general support the automatic evaluation findings. Yet some score, show need sent eval

% \subsection{}
\subsection{Application}

% \noindent\textbf{RQ: Is the model able to generate contrasting opinions on controversial issues }.
We hypothesize that the formulation of the task enables the application of discovering groups with different opinions on issues. We verify the hypothesis by collecting personas with contrasting stances on an issue and generating responses based on this issue. We find that the output from the model stays consistent with the persona (examples are shown in the Appendix Table~\ref{tab:contrast}). The result demonstrates the potential for application on social network analysis. Since the model is able to generalize to different personas or news, an analyst can therefore replace the news headline with others to segment the population based on different issues, or manually construct a persona to visualize how a person from a particular community would respond to certain issues. %~\ref{tab:contrast}

% \label{sec:humaneval}
% \subsection{}
% Shown in Table ...
% \subsection{Analysis of Lurkers}

% \subsection{Error Analysis}

\section{Conclusions and Future Work}
\label{sec:conclusion}

%In this work, 
We propose Response Forecasting on Personas for News Media, a new task that tests the model's capability of estimating the responses from different personas. The task enables important applications such as estimating the effect of unreleased messages on different communities as an additional layer of defense against unsafe information (e.g., information that might cause conflict or moral injury). We also create the first dataset for evaluating this new task and present an evaluation of the state-of-the-art neural models. The empirical results show that the best-performing models are able to predict responses with reasonable accuracy and produce outputs that are consistent with the personas. The analysis shows that the models are also able to generate contrasting opinions when conditioned on contrasting personas, demonstrating the feasibility of applying the models to discovering social groups with different opinions on issues for future work. In addition to this, an intriguing avenue for further research lies in utilizing response forecasting techniques to predict the popularity of discussion threads, as explored in previous studies~\cite{DBLP:conf/emnlp/HeOHCGLD16,DBLP:conf/emnlp/ChanK18}.

\section*{Limitations}
\label{sec:limitation}

While the training method makes use of user profile description and history, one additional factor that is important is the structure between users and news articles. Knowing a user's social circles can often give hints about the user's interests and beliefs, which can potentially help the model to infer how a particular persona would respond to an issue. A possible direction is to design a method that explores the social context features (e.g., social network) via graph-based algorithms. 
% Meanwhile, more categories can be defined to help evaluate response prediction; for instance, we can measure how the beliefs of a user would change after seeing a particular news message.

% While TCD paired with concept acquisition methods can aid downstream script learning tasks, it doesn't consider the inclusion of actions of each step event, which can potentially benefit the script learning tasks. A possible direction is to extend the design of TCD and the concept prompt to include the semantics of actions and their orders.  

% Meanwhile, the concepts extracted by our method do not overlap with the ground-truth concepts (i.e., the set of concepts that appear in the reference) very well (e.g., <20\% in Jaccard Index). The gap in performance between our methods and the Gold-Concept variant shows that improving the concept derivation quality might be the next step.

% Furthermore, because our dataset is constructed from the English version of WikiHow, the benefits of our methods shown in the experiments are only empirically proved to work for English. We plan to further test our methods in multiple languages.

\section*{Ethics}\label{sec:ethics} 
During annotation, each worker was paid \$15 per hour (converted to per assignment cost on MTurk). If workers emailed us with any concerns, we responded to them within 1 hour. The research study has also been approved by the Institutional Review Board (IRB) and Ethics Review Board at the researchers’ institution. Regarding privacy concerns our dataset may bring about, we follow the Twitter API’s Terms of Use\footnote{\url{https://developer.twitter.com/en/developer-terms/agreement-and-policy}} and only redistribute content for non-commercial academic research only. We will release pointers to the tweets and user profiles in the dataset.
% We will additionally require interested parties to sign an agreement on data and model usage to make sure the resource will be used in ethical ways and available for academic use only.

% Individuals redistributing Tweet IDs and/or User IDs on behalf of an academic institution for the sole purpose of non-commercial research are permitted to redistribute an unlimited number of Tweet IDs and/or User IDs.

% Considering the privacy violation problems our dataset may bring about, we followed Reddit’s term of use for user content—based on Reddit API Terms of Use, users are granted with license to display the user content through application9.

% We followed the guidelines for ethical annotation practices and crowdsourcing that are outlined in (Sheehan, 2018), including paying workers a fair wage above the federal minimum. If workers contacted us with any questions or concerns, we responded promptly to them within 24 hours. In the task interface, in the header, we warned annotators that the content might be upsetting, and we gave the following recommendation: “if any point you do not feel comfortable, please feel free to skip the HIT or take a break.”.

% Scientific work published at ACL 2023 must comply with the ACL Ethics Policy.\footnote{\url{https://www.aclweb.org/portal/content/acl-code-ethics}} We encourage all authors to include an explicit ethics statement on the broader impact of the work, or other ethical considerations after the conclusion but before the references. The ethics statement will not count toward the page limit (8 pages for long, 4 pages for short papers).

\section*{Acknowledgement} This research is based upon work supported in part by U.S. DARPA INCAS Program No. HR001121C0165. The views and conclusions contained herein are those of the authors and should not be interpreted as necessarily representing the official policies, either expressed or implied, of DARPA, or the U.S. Government. The U.S. Government is authorized to reproduce and distribute reprints for governmental purposes notwithstanding any copyright annotation therein. Hou Pong Chan was supported in part by the Science and Technology Development Fund, Macau SAR (Grant Nos. FDCT/060/2022/AFJ, FDCT/0070/2022/AMJ) and the Multi-year Research Grant from the University of Macau (Grant No. MYRG2020-00054-FST).

\bibliography{anthology,custom}
\bibliographystyle{acl_natbib}
% \section*{Acknowledgements}
% This document has been adapted by Jordan Boyd-Graber, Naoaki Okazaki, Anna Rogers from the style files used for earlier ACL, EMNLP and NAACL proceedings, including those for
% EACL 2023 by Isabelle Augenstein and Andreas Vlachos,
% EMNLP 2022 by Yue Zhang, Ryan Cotterell and Lea Frermann,
% ACL 2020 by Steven Bethard, Ryan Cotterell and Rui Yan,
% ACL 2019 by Douwe Kiela and Ivan Vuli\'{c},
% NAACL 2019 by Stephanie Lukin and Alla Roskovskaya, 
% ACL 2018 by Shay Cohen, Kevin Gimpel, and Wei Lu, 
% NAACL 2018 by Margaret Mitchell and Stephanie Lukin,
% Bib\TeX{} suggestions for (NA)ACL 2017/2018 from Jason Eisner,
% ACL 2017 by Dan Gildea and Min-Yen Kan, NAACL 2017 by Margaret Mitchell, 
% ACL 2012 by Maggie Li and Michael White, 
% ACL 2010 by Jing-Shin Chang and Philipp Koehn, 
% ACL 2008 by Johanna D. Moore, Simone Teufel, James Allan, and Sadaoki Furui, 
% ACL 2005 by Hwee Tou Ng and Kemal Oflazer, 
% ACL 2002 by Eugene Charniak and Dekang Lin, 
% and earlier ACL and EACL formats written by several people, including
% John Chen, Henry S. Thompson and Donald Walker.
% Additional elements were taken from the formatting instructions of the \emph{International Joint Conference on Artificial Intelligence} and the \emph{Conference on Computer Vision and Pattern Recognition}.

% % Entries for the entire Anthology, followed by custom entries
% \bibliographystyle{acl_natbib}
\appendix

% \pagebreak
% \pagebreak

\section{Appendix}
\label{sec:appendix}

\subsection{Implementation Details}
\label{app:trainig}

We implement the models using the 4.8.2 version of Huggingface Transformer library\footnote{\url{https://github.com/huggingface/transformers}}\cite{huggingface}. We use Oct 1, 2021 commit version of the BART-base model (139M parameters) from Huggingface\footnote{\url{https://huggingface.co/facebook/bart-base/commit/ea0107eec489da9597e9eefd095eb691fcc7b4f9}}. We use Huggingface datasets\footnote{\url{https://github.com/huggingface/datasets}} for automatic evaluation metrics. The BART Score comes from the author's repository\footnote{\url{https://github.com/neulab/BARTScore}} and we used the one trained on ParaBank2. The hyperparameters for the experiment are shown in Table~\ref{tab:hparam} (applied to all models) and the ones not listed in the table are set to be default values from the transformer library. In order to make the distribution of training and development sets align, we used automatically-generated labels\footnote{\url{https://huggingface.co/cardiffnlp/twitter-roberta-base-sentiment-latest},\url{https://competitions.codalab.org/competitions/17751}} during training. We use RAdam~\cite{radam} as the optimizer. We perform hyperparameter search on the batch size from \{16, 32\}, pretrained language model learning rate from \{3e-5, 4e-5, 5e-5\}. We perform our experiments on 32 GB V100. The experiments can take up to 15 hours.

\begin{table}[]
	\centering
 % \begin{small}
	\begin{tabular}{ll}
		
		\toprule
		Name &Value\\
		\midrule
		seed &42  \\
		learning rate & 5e-5  \\
		batch size & 16  \\
		weight decay & 5e-4  \\
		RAdam epsilon & 1e-8  \\
		RAdam betas & (0.9, 0.999)  \\
		scheduler & linear  \\
		warmup ratio (for scheduler) & 0.06  \\
		number of epochs & 20  \\
		metric for early stop & SacreBLEU\footnote{\url{https://github.com/mjpost/sacrebleu}} \\
		patience (for early stop) & 15  \\ \midrule
		length penalty & 1.2  \\
		% max length & 511  \\
		% min length & 2  \\
		{beam search size during eval} & 5  \\
		\bottomrule
	\end{tabular}
 % \end{small}
 \caption{Hyperparameters. The ones below the mid-line are generation related.}
	\label{tab:hparam}
\end{table}

% \usepackage{multirow}
% \usepackage[normalem]{ulem}
% \usepackage{multirow}
% \usepackage[normalem]{ulem}
% \usepackage{multirow, makecell}
% \usepackage{multirow}
% \begin{table*}[]
% \begin{small}
% \centering
% \begin{tabular}{|c|c|c|}
%  \toprule
%  \hline
%  Text P  & Text P    & Text Q     \\\hline
%  \multirow{2}*{\shortstack[l]{Mllions are under a blizzard warning \\as a powerful storm is expected\\ to bring heavy snow,\\ wind and rain to a large swath \\of the country}} & xt P    & Text Q     \\ \cline{2-3}
%      &           Text P     & Text P         \\
% \bottomrule
% \end{tabular}
% \end{small}
% \end{table*}

\begin{table}[t]
% \begin{small}
\centering
\begin{tabular}{lll}
\toprule
\multicolumn{2}{l}{\textbf{\begin{tabular}[c]{@{}l@{}}Headline: Millions are under a blizzard\\  warning as a powerful storm is expected to\\  bring heavy snow, wind and rain to a \\large  swath of the country\end{tabular}}} \\ \midrule
\multicolumn{1}{l|}{Purity \& Love} & Degradation \\ \midrule
\multicolumn{1}{l|}{\begin{tabular}[c]{@{}l@{}}We’re in the northern \\ part of the country. \\ Hope everyone is safe\end{tabular}} & \begin{tabular}[c]{@{}l@{}}Mother Nature sure \\ is pissed off at us\end{tabular} \\ 
\bottomrule
\end{tabular}

\begin{tabular}{lll}
\toprule
\multicolumn{2}{l}{\textbf{\begin{tabular}[c]{@{}l@{}}Headline: Judge says Trump may have been \\urging supporters to 'do something more' \\ than protest on Jan. 6\end{tabular}}} \\ \midrule
\multicolumn{1}{l|}{Pro-President Trump} & {Anti-President Trump }\\ \midrule
\multicolumn{1}{l|}{\begin{tabular}[c]{@{}l@{}}The liberal media \& \\Dems are always \\negative when it \\comes to anything.\\ They don't care \\about anything \\except themselves\end{tabular}} & \begin{tabular}[c]{@{}l@{}} Hahahahahahaha! \\They figured that \\Trump would be \\impeachedby now!  \\But the traitorous\\ Republicans are \\slowing down the \\process.\end{tabular} \\ 
\bottomrule
\end{tabular}

\begin{tabular}{lll}
\toprule
\multicolumn{2}{l}{\textbf{\begin{tabular}[c]{@{}l@{}}Headline: Russia and Ukraine are at war
\end{tabular}}} \\ \midrule
\multicolumn{1}{l|}{Pro-Russia} & Pro-Ukraine \\ \midrule
\multicolumn{1}{l|}{\begin{tabular}[c]{@{}l@{}}Support Russia \end{tabular}} & \begin{tabular}[c]{@{}l@{}}Support Ukraine\end{tabular} \\ 
\bottomrule
\end{tabular}

\caption{Tables showing different cases that contrasting the persona (selected from existing ones) can lead to the generation of contrasting opinions on issues. For each table, the middle row contains different personas, and the third row contains the responses from each persona.}
\label{tab:contrast}
% \end{small}
\end{table}

\end{document}